\documentclass[letterpaper, 10 pt, conference, twoside]{IEEEtran} 
\usepackage{graphicx}
\graphicspath{./}	
\DeclareGraphicsExtensions{.pdf,.jpeg,.png}
\usepackage{color}
\usepackage{subfig}
\usepackage{mathptmx}   
\usepackage{amssymb}                        
\usepackage{amsmath}                            
\usepackage{amsfonts}
\usepackage{mathtools}
\usepackage{algpseudocode}
\usepackage{algorithm}
\usepackage{tabularx}
\usepackage{booktabs}
\usepackage{colortbl} 
\usepackage{xcolor} 
\usepackage{xfrac}
\usepackage{multirow}
\usepackage{bm}
\usepackage{amsbsy}
\usepackage{fixmath}
\usepackage{lettrine}

\title{
SATR-DL: Improving Surgical Skill Assessment and Task Recognition in Robot-assisted Surgery with Deep Neural Networks
}

\author{Ziheng~Wang,~\IEEEmembership{Student Member, IEEE}
 Ann~Majewicz Fey~\IEEEmembership{Member, IEEE}
}%

\begin{document}
\maketitle

\begin{abstract}
\noindent \textit{Purpose:} 
This paper focuses on an automated analysis of surgical motion profiles for objective skill assessment and task recognition in robot-assisted surgery. Existing techniques heavily rely on conventional statistic measures or shallow modelings based on hand-engineered features and gesture segmentation. Such developments require significant expert knowledge, are prone to errors, and are less efficient in online adaptive training systems. 
\noindent \textit{Methods:} 
In this work, we present an efficient analytic framework with a parallel deep learning architecture, SATR-DL, to assess trainee expertise and recognize surgical training activity. Through an end-to-end learning technique, abstract information of spatial representations and temporal dynamics is jointly obtained directly from raw motion sequences. 
\noindent \textit{Results:} 
By leveraging a shared high-level representation learning, the resulting model is successful in the recognition of trainee skills and surgical tasks, \textit{suturing}, \textit{needle-passing}, and \textit{knot-tying}.
Meanwhile, we explore the use of ensemble in classification at the trial level, where the SATR-DL outperforms state-of-the-art performance by achieving accuracies of 0.960 and 1.000 in skill assessment and task recognition, respectively.  
\noindent \textit{Conclusion:} 
This study highlights the potential of SATR-DL to provide improvements for an efficient data-driven assessment in intelligent robotic surgery.
\end{abstract}

\section{INTRODUCTION}

 \begin{figure*}[!h]
      \centering
     \includegraphics [width=0.99\linewidth,clip,trim=0pt 10pt 0pt 20pt]{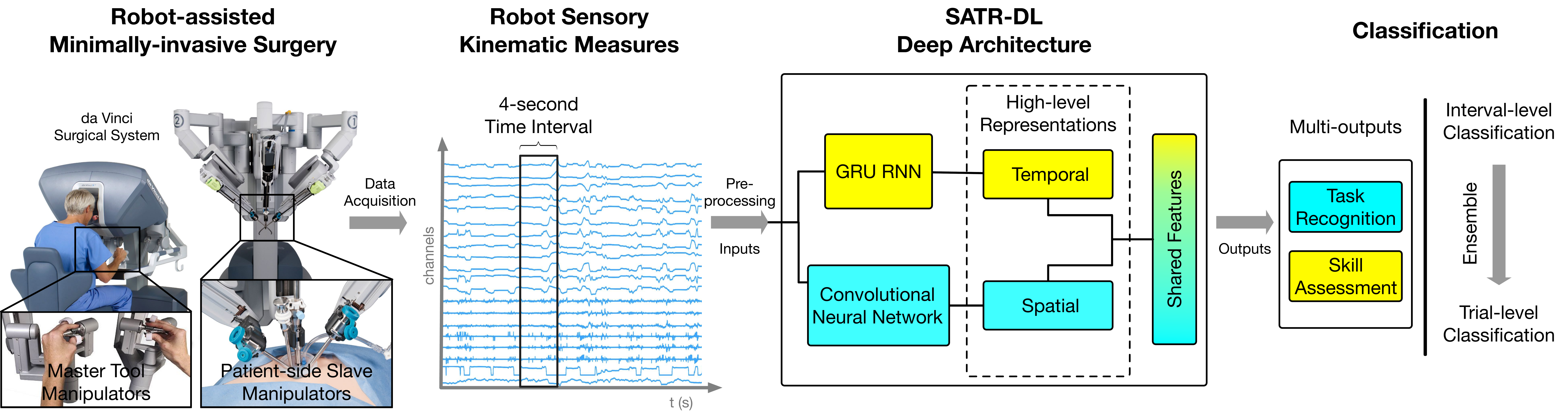}
	\caption{ {\bf An end-to-end framework for online skill analysis and recognition in robot-assisted minimally-invasive surgery.} Data flow is from left to right. The framework takes 4-second interval of motion data as input, recorded from \textit{daVinci} end-effectors. By learning raw sequences spatially and temporally in parallel, it outputs discriminative recognition of trainee skills and training tasks at the level of interval frame (once every 4 seconds). For trial-level assessment, classification is obtained by the majority-vote via ensemble.} 
      \label{fig: framework} 
     \vspace{-0.2cm}
\end{figure*}

In robot-assisted minimally-invasive surgery, surgeon expertise directly affects overall surgical performance and patient safety. Technical training programs are necessary to ensure trainees develop adequate skills to teleoperate robots and perform complex operations proficiently. 
Due to the steep learning curves, an objective measure of trainee performance and automated identification of surgical activity are of prominent concerns towards an efficient training and intelligent robot autonomy in order to further enhance surgery outcomes~\cite{reiley2011objective,kassahun2016surgical}.  

Current techniques for objective surgical skill assessment include descriptive statistics (time, path length, smoothness, etc.)~\cite{nisky2013effect,ershad2016meaningful}, gesture segmentation-based analysis such as Hidden Markov Models (HMM)~\cite{tao2012sparse}, and feature-based modelings such as k-nearest neighbor ($k$NN), support vector machine (SVM)~\cite{forestier2017jigsaw}. 
Given an observation of motion data from robot end-effectors, local segmented gestures or hand-crafted features are extracted and fed into a classifier to assess trainee skills and performance.
Similarly, these techniques are applied to understand underlying surgical task structures and workflow~\cite{reiley2008automatic}. However, the aforementioned approaches are limited in several ways. First, it is time-consuming and strenuous to manually design meaningful representations to uncover hidden pattens of complex motion. Gesture segmentation is task-dependent, limited to specific operations and requires significant prior knowledge of particular structures and pre-processing to decompose motion sequences. Moreover, a common deficiency is that most classifications can only obtained at the level of trial, which requires an entire observation of each training operation. These drawbacks make previous approaches less efficient for an online automatic feedback system. A sequence distance-based method, such as dynamic time warping (DTW), has been proposed to provide feasible online classifying for surgical task and gesture recognition~\cite{fard2017distance}. However, higher computational loads involved in practice, as well as the role of DTW in the skill analysis still remains unknown.
 
In this paper, we aim at developing an end-to-end surgery motion analytic framework based on a multi-output deep learning architecture, SATR-DL, for online trainee skill analysis and task recognition (Figure~\ref{fig: framework}). In particular, by integrating a Convolutional Neural Network (CNN)~\cite{szegedy2015going} and Gated Recurrent Unit (GRU) network~\cite{chung2014empirical}, our proposed deep model can simultaneously learn both spatial (convolutional) abstract representations within the interval of input frame, as well as the temporal dynamics of multiple channels at each time step in raw motion data.
By exploring these intrinsic properties of kinematic sequences, SATR-DL can effectively characterize the nature of surgery motion relative to both the trainee experience and operation activity. 
Crucially, this work does not assume any prior knowledge of primitive gestures and does not require pre-defined features.
We find that our SATR-DL significantly enhances both the efficiency and accuracy when compared to other techniques in the study of robotic surgery assessment.   
  
\section{METHODOLOGY}
\subsection{Problem Formulation}

The goal of the SATR-DL model is to directly map raw motion kinematics onto labels of trainee skills and corresponding training tasks. This can be formalized as a supervised classification problem with multiple outputs, given a time interval of high-dimension motion sequences as input, $X\in\mathbb{R}^{T\times C}$. Here, $T$ refers to the time steps in each interval frame and $C$ is the number of channels of the input. The outputs are label $y_1 \in \{1 :``novice", 2:``intermediate'', 3 :``expert" \}$ for skill levels, and label $y_2 \in \{1 :``suturing", 2:``\textit{needle-passing}'', 3 :``\textit{knot-tying}" \}$ for training tasks. 
We take the jointed cross-entropy loss as the global objective function (Eq.~\ref{eq: cross-entropy}), which measures the total discrepancy between the ground-truth and predicted labels of two outputs.

\vspace{-0.4cm}

\begin{equation}
\label{eq: cross-entropy}
J(\theta) = -\displaystyle\sum_{j=1}^{l} \displaystyle\sum_{i=1}^{m}\displaystyle\sum_{k=1}^{K}{ 1 \{y_{j}^{(i)}=k \}  \log{p(y_{j}^{(i)}=k | x^{(i)}; \theta )} }
\end{equation} 
where $l$ is the total output number ($l=2$ for skill and task classification), $m$ is the number of training examples, $K$ is the total label number for each output ($K=3$), and $p(y_{j}^{(i)}=k | x^{(i)}; \theta)$ is the conditional likelihood that the predicted label $y_{j}^{(i)}$ of output $j$ on a single example $x^{(i)}$ assigns to the $k$-th label, given the trained model parameters $\theta$.   

The classification is conducted for each time interval frame (every 4 seconds). In addition, for trial-level assessment, a majority voting is used on the whole trial sequence based on the ensemble of interval-level classification in order to deliver a robust classification performance. 


\subsection{Overall Architecture}

\begin{figure*}[tb]
      \centering
     \includegraphics [width=0.85\linewidth,clip,trim=10pt 10pt 0pt 20pt]{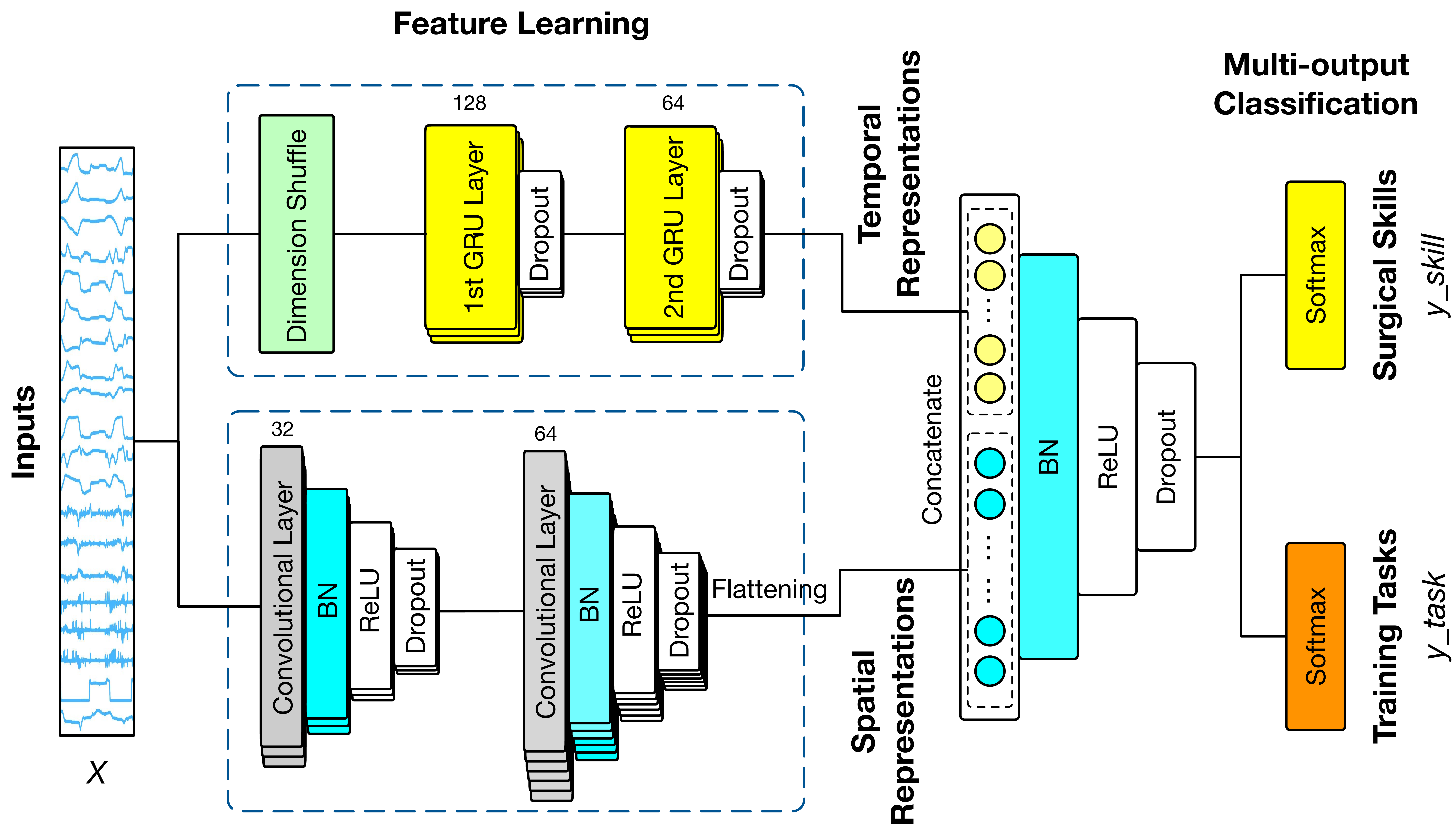}
     \caption{ {\bf Proposed SATR-DL architecture with multi-output classifications.} The model utilizes convolutional and GRU recurrent neural networks via parallel computing to learn both spatial and temporal representations from low to high levels. 
		} 
      \label{fig: architecture} 
\vspace{-0.5cm}
\end{figure*}

Figure~\ref{fig: architecture} shows the design of our SATR-DL model. The proposed architecture is composed of a temporal component that decodes temporal information across each time step and a convolutional component that explores implicit representations over spatial (convolutional) dimensions in the input frame. 
Specifically, the convolutional component consists of two stacked basic blocks with kernel numbers of 32 and 64, respectively.
Each block consists of a convolution layer with 1-D kernel of size 2, followed by a batch normalization (\textit{BN})~\cite{ioffe2015batch}, a rectified linear unit (\textit{ReLU}) activation~\cite{nair2010relu}, and a dropout layer (dropout rate 0.2)~\cite{srivastava2014dropout}. 
In parallel, two stacked GRU recurrent neural networks are applied to process the input for learning temporal dynamics of motion sequences. The hidden units for the first and second GRU layers are set as 128 and 64, respectively. A dropout with the probability of 0.2 is applied after each GRU layer. Next, the output of convolutional block with flattening and the 2-layer GRU are reconstructed by a concatenation component, followed by a batch normalization, \textit{ReLU} activation and a dropout with the rate of $0.5$. Dropping hidden units at higher layers forces the network to learn more compact and abstract representations and reduce overfitting. Finally, a multi-output classification is obtained from two separated softmax logistic regressions for classifying skill and training tasks.

\section{EXPERIMENTAL EVALUATION}
\subsection{Dataset}
We test the proposed SATR-DL on the public-available minimally invasive surgical dataset, JHU-ISI Gesture and Skill Assessment Working Set (JIGSAWS). Details about data collection and variables are described in \cite{gao2014JIGSAW}. The kinematic motion is captured as multi-channel time-series data from \textit{daVinci} end-effectors with 30 Hz sampling frequency. Four \textit{novice} trainees (who practice on the \textit{dVSS} $<10$ hours), two \textit{intermediate} trainees ($10-100$ hours practice) and two \textit{expert} surgeons ($>100$ hours practice) performed five repetitions each of three standard surgical training tasks, \textit{suturing}, \textit{needle-passing}, and \textit{knot-tying}.

\subsection{Pre-processing}
Raw motion sequences are first normalized to a zero mean with unit variance for each individual sensory channel. The original sequences are further processed with a sliding window of size 120 (4-second record) with a step size of 30, yielding a compatible format for the network input regardless of varied recording durations, since the time sequence may vary for each trial. This approach also enables us to obtain a large set of examples applicable for deep learning to avoid substantial overfitting. 

\subsection{Training \& Performance Validation}
In this work, all parameters of the network layers are initialized using the Xavier initialization~\cite{2015DL_review}.
We train the SATR-DL by minimizing the global cross-entropy loss function using an adaptive learning method. 
Network parameters in processing units are jointly optimized using mini-batch gradient descent for training efficiency with the Adam update rule~\cite{kingma2014adam}. Using this configuration, a total of 80 epochs with 600 mini-batches is run for training. The initial learning rate, $\epsilon$, is set as 0.005, and the exponential decay rates of the first and second moment estimates are 0.9 and 0.999, respectively. The learning rate is reduced by a factor of 5 once the overall validation loss has stopped improving every three epochs. To obtain a robust learning model, the above hyper-parameters are fine-tuned via evaluation on validation set, which is split from the training data with the $80/20$ partition for training and validation. The best model is chosen from the one that achieves the highest accuracy in validation set.

To validate our proposed deep architecture, we adopt the \textit{Leave-One-Supertrial-Out} cross-validation scheme (\textit{LOSO}), which is widely used in current skill assessment studies~\cite{gao2014JIGSAW}. 

\section{RESULTS AND DISCUSSION}
We evaluate the classification performance regarding the \textit{precision}, \textit{recall}, and \textit{f1-score} for each output class, and the overall \textit{accuracy}. These performance metrics are reported as averages of classification across all five-fold \textit{LOSO} cross-validation evaluated on each testing set. 

\begin{figure}[tb]
      \centering
      \includegraphics [width=0.90\linewidth,clip,trim=5pt 5pt 5pt 10pt]{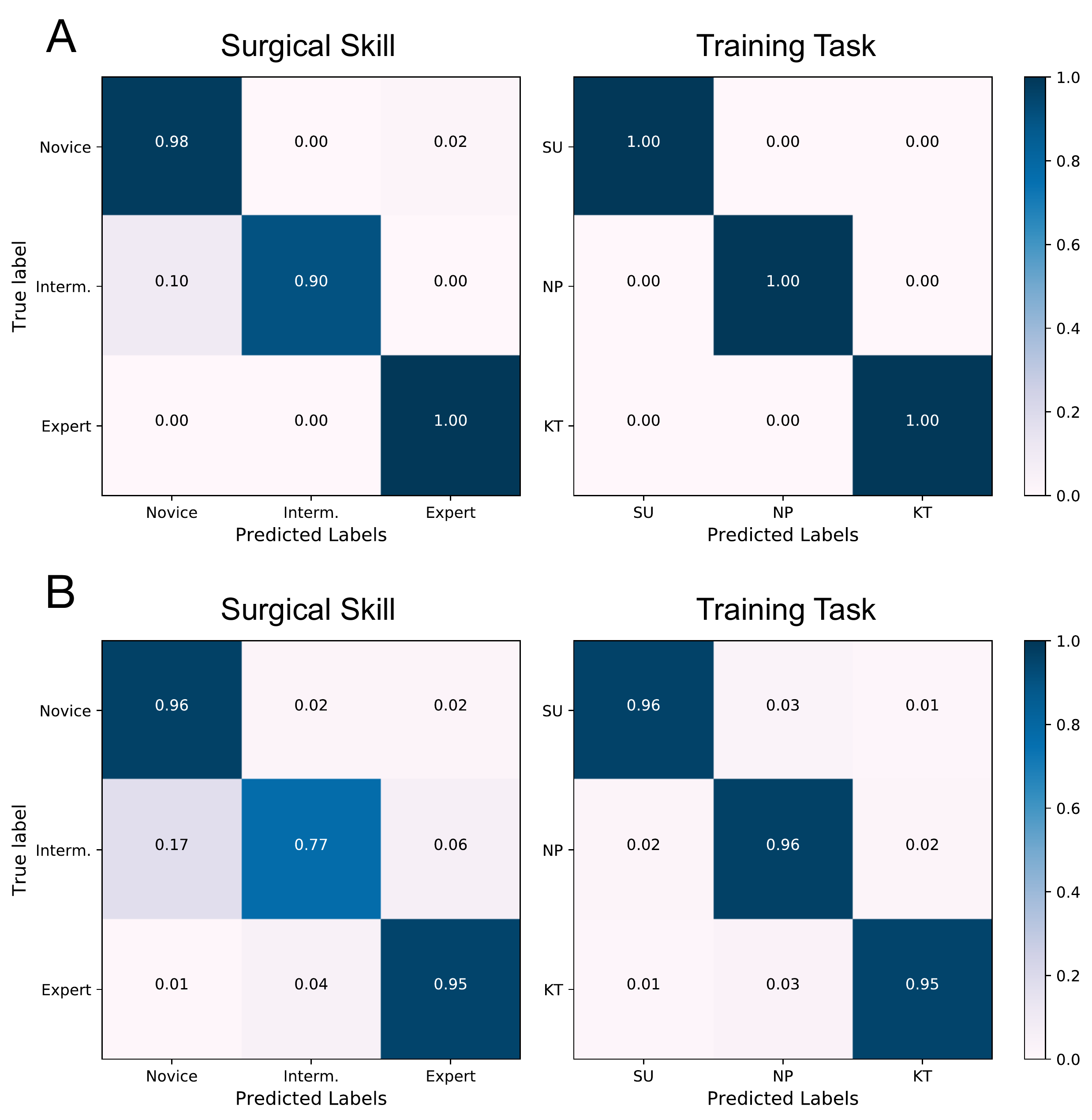}
      \caption{ {\bf 
      Confusion matrices of SATR-DL results in five-fold \textit{LOSO} cross-validation.} (A) trial-level classification, (B) interval-level classification with 4-second kinematics input. 
      }
      \label{fig: confusion}
\vspace{-0.5cm}
\end{figure}

Figure~\ref{fig: confusion} shows the normalized confusion matrices of SATR-DL predictions for each output class. Table~\ref{tab: rts_perform} gives a summary of model performance in the skill and task recognition. 
This model correctly assesses the trainee expertise with an overall accuracy of 0.920, given a 4-second motion sequence input. 
Note that a higher accuracy of 0.960 in the recognition of training tasks is achieved compared to surgical skill levels. This result might potentially benefit from the learned information of temporal dynamics, since different tasks are associated with varied motion patterns in the temporal dimension.
Table~\ref{tab: rts_perform} also compares the aggregated results of classification at the level of trial compared to the interval-level assessment. This result provides evidence that the SATR-DL algorithm can yield better accuracy using ensemble of interval-level classification for each trial.  

Furthermore, we compare the overall performance of SATR-DL model with that of state-of-the-art methods in both surgical skill assessment and training recognition, as shown in Table~\ref{tab: rts_comparison}. All benchmarks are conducted under \textit{LOSO} cross-validation scheme based on JIGSAWS kinematics data. Using a 4-second data window input, our model outperforms conventional feature-based methods including $k$NN, LR, and SVM, with at least $6.85\%$ improvements in trainee skill assessment, except for the sparse HMM (S-HMM). For task recognition, SATR-DL provides at least $1.15\%$ accuracy improvements compared with existing methods of HMM, DTW-$k$NN. Also, for classification at the level of trial, SATR-DL achieves the best performance compared to all other methods, with its highest accuracy of 0.960 in skill assessment and 1.000 in task recognition, respectively. 

\section{CONCLUSION}
In this work, we propose a novel analytical framework with deep feature learning, SATR-DL, to investigate multi-channel motion kinematic profiles for an automatic surgical skill assessment and task recognition. The benefits of our proposed SATR-DL are: 1) highly-efficient processing of raw surgery motion via end-to-end learning given a limited data influx, and 2) improved performance of assessment in recognizing trainee skills and surgical activities.
Combining a CNN with GRU component, we demonstrate that the SATR-DL architecture can jointly exploit the intrinsic information spatially and sequentially from high-dimensional surgical motion data. In addition to interval-level assessment, we also show that the classification at the level of each trial can exceed published results, derived from conventional statistics and machine learning methods, with an improvement of at least 6.85\% and 1.15\% in skill analysis and task recognition, respectively.
These deep architectures and high-level motion analysis techniques could have the potential to largely improve trainee assessment for robotic minimally invasive surgery. 

\begin{table*}[tb]
\centering
\caption{ {\bf Summary table of SATR-DL classification performance employed for skill assessment and task recognition at interval-level and trial-level.} Performance results on testing sets under \textit{LOSO} validation scheme are reported in terms of \textit{precision}, \textit{recall}, \textit{f1-score} for each class, and overall \textit{accuracy}.  
}
\label{tab: rts_perform}
\renewcommand\arraystretch{1.2}
\renewcommand\tabcolsep{11.2pt}
\begin{tabular}{llllllllll}
\toprule[\heavyrulewidth]         
\multicolumn{2}{l}{\multirow{2}{*}{\textbf{  }}} & \multicolumn{4}{l}{\textbf{Interval-level Classification}} & \multicolumn{4}{l}{\textbf{Trial-level Classification}} \\ \cmidrule(lr){3-6} \cmidrule(lr){7-10}
\multicolumn{2}{l}{} & \textit{precision} & \textit{recall} & \textit{f1-score} & \parbox{1.05cm}{\textit{overall accuracy}}  & \textit{precision} & \textit{recall} & \textit{f1-score} & \parbox{1.05cm}{\textit{overall accuracy}} \\ 
\midrule[\heavyrulewidth]     
\multirow{3}{*}{\parbox{0.8cm}{\textbf{Surgical Skill}}} & \textit{Novice} &  0.94 & 0.96 & 0.95 & \multirow{3}{*}{0.920}  & 0.95 &  0.98 & 0.97 &   \multirow{3}{*}{0.966}  \\
 & \textit{Intermediate} & 0.88 & 0.77 & 0.82 &  & 1.00 & 0.90 & 0.95 &  \\
 & \textit{Expert} & 0.90 & 0.95 & 0.93 &  & 0.97 &  1.00 & 0.98 &  \\
 \cmidrule(lr){1-10}    
\multirow{3}{*}{\parbox{0.8cm}{\textbf{Training Task}}} & \textit{Suturing} & 0.97 & 0.96 & 0.97 &  \multirow{3}{*}{0.960}  & 1.00 & 1.00 & 1.00 &   \multirow{3}{*}{1.000} \\
 & \textit{Needle-passing} & 0.95 & 0.96 & 0.96 &  & 1.00 & 1.00 & 1.00 &  \\
 & \textit{Knot-tying} & 0.95 & 0.95 & 0.95 &  & 1.00 & 1.00 & 1.00  &  \\
 \midrule[\heavyrulewidth]     
\end{tabular}
\vspace{-0.5cm}
\end{table*}

\begin{table}[tb]
\centering
\caption{ {\bf Performance comparison of our proposed SATR-DL with state-of-the-art results under \textit{LOSO} validation in terms of overall \textit{accuracy}.} The best classification results are highlighted in bold.  
}
\label{tab: rts_comparison}
\renewcommand\arraystretch{1.05}
\renewcommand\tabcolsep{5pt}
\begin{tabular}{llcc}
\toprule[\heavyrulewidth]     
\multirow{2}{*}{\textbf{Author, \textit{Year}}} & \multirow{2}{*}{\textbf{Method}} & \multicolumn{2}{c}{\textbf{\textit{Overall Accuracy}}}  \\  \cmidrule(lr){3-4}
 &  & \textbf{Skill Assessment} & \textbf{Task Recognition}   \\
 \midrule[\heavyrulewidth]     
 \multirow{2}{*}{\parbox{1.8cm}{Lingling \emph{et al.}, \\\textit{2012}~\cite{tao2012sparse}}}   & \multirow{2}{*}{S-HMM} & \multirow{2}{*}{0.960} & \multirow{2}{*}{$-$} \\
 &  &  &   \\
\cmidrule(lr){1-4}
\multirow{3}{*}{\parbox{1.8cm}{Fard \emph{et al.}, \\\textit{2017}~\cite{forestier2017jigsaw}}}   & $k$NN & 0.859 & \multirow{3}{*}{$-$} \\
 &  LR &  0.861 &    \\
 & SVM  & 0.754 &  \\
 \cmidrule(lr){1-4}
 \multirow{2}{*}{\parbox{1.8cm}{Fard \emph{et al.}, \\\textit{2017}~\cite{fard2017distance}}}   & HMM & \multirow{2}{*}{$-$}  &   0.924 \\
 &  DTW-$k$NN &  & 0.955  \\
  \cmidrule(lr){1-4}
 \multirow{2}{*}{\parbox{1.8cm}{Current study\\SATR-DL}}   & \multirow{2}{*}{CNN-GRU} & 0.920 &   0.966 \\ \cmidrule(lr){3-4}
 &  & \textbf{0.960}  & \textbf{1.000} \\
 \midrule[\heavyrulewidth]     
\end{tabular}
\vspace{-0.5cm}
\end{table}


\bibliographystyle{IEEEtran}

\begin{thebibliography}{10}
\providecommand{\url}[1]{#1}
\csname url@samestyle\endcsname
\providecommand{\newblock}{\relax}
\providecommand{\bibinfo}[2]{#2}
\providecommand{\BIBentrySTDinterwordspacing}{\spaceskip=0pt\relax}
\providecommand{\BIBentryALTinterwordstretchfactor}{4}
\providecommand{\BIBentryALTinterwordspacing}{\spaceskip=\fontdimen2\font plus
\BIBentryALTinterwordstretchfactor\fontdimen3\font minus
  \fontdimen4\font\relax}
\providecommand{\BIBforeignlanguage}[2]{{%
\expandafter\ifx\csname l@#1\endcsname\relax
\typeout{** WARNING: IEEEtran.bst: No hyphenation pattern has been}%
\typeout{** loaded for the language `#1'. Using the pattern for}%
\typeout{** the default language instead.}%
\else
\language=\csname l@#1\endcsname
\fi
#2}}
\providecommand{\BIBdecl}{\relax}
\BIBdecl

\bibitem{reiley2011objective}
C.~E. Reiley, H.~C. Lin, D.~D. Yuh, and G.~D. Hager, ``Review of methods for
  objective surgical skill evaluation,'' \emph{Surgical Endoscopy}, vol.~25,
  no.~2, pp. 356--366, 2011.

\bibitem{kassahun2016surgical}
Y.~Kassahun, B.~Yu, A.~T. Tibebu, D.~Stoyanov, S.~Giannarou, J.~H. Metzen, and
  E.~Vander~Poorten, ``Surgical robotics beyond enhanced dexterity
  instrumentation: a survey of machine learning techniques and their role in
  intelligent and autonomous surgical actions,'' \emph{International Journal of
  Computer Assisted Radiology and Surgery}, vol.~11, no.~4, pp. 553--568, 2016.

\bibitem{nisky2013effect}
I.~Nisky, M.~H. Hsieh, and A.~M. Okamura, ``The effect of a robot-assisted
  surgical system on the kinematics of user movements,'' in \emph{Engineering
  in Medicine and Biology Society (EMBC), 2013 35th Annual International
  Conference of the IEEE}.\hskip 1em plus 0.5em minus 0.4em\relax IEEE, 2013,
  pp. 6257--6260.

\bibitem{ershad2016meaningful}
M.~Ershad, Z.~Koesters, R.~Rege, and A.~Majewicz, ``Meaningful assessment of
  surgical expertise: Semantic labeling with data and crowds,'' in
  \emph{International Conference on Medical Image Computing and
  Computer-Assisted Intervention}.\hskip 1em plus 0.5em minus 0.4em\relax
  Springer, 2016, pp. 508--515.

\bibitem{tao2012sparse}
L.~Tao, E.~Elhamifar, S.~Khudanpur, G.~D. Hager, and R.~Vidal, ``Sparse hidden
  markov models for surgical gesture classification and skill evaluation.'' in
  \emph{IPCAI}.\hskip 1em plus 0.5em minus 0.4em\relax Springer, 2012, pp.
  167--177.

\bibitem{forestier2017jigsaw}
G.~Forestier, F.~Petitjean, P.~Senin, F.~Despinoy, and P.~Jannin, ``Discovering
  discriminative and interpretable patterns for surgical motion analysis,'' in
  \emph{Conference on Artificial Intelligence in Medicine in Europe}.\hskip 1em
  plus 0.5em minus 0.4em\relax Springer, 2017, pp. 136--145.

\bibitem{reiley2008automatic}
C.~E. Reiley, H.~C. Lin, B.~Varadarajan, B.~Vagvolgyi, S.~Khudanpur, D.~Yuh,
  and G.~Hager, ``Automatic recognition of surgical motions using statistical
  modeling for capturing variability,'' \emph{Studies in health technology and
  informatics}, vol. 132, p. 396, 2008.

\bibitem{fard2017distance}
M.~J. Fard, A.~K. Pandya, R.~B. Chinnam, M.~D. Klein, and R.~D. Ellis,
  ``Distance-based time series classification approach for task recognition
  with application in surgical robot autonomy,'' \emph{The International
  Journal of Medical Robotics and Computer Assisted Surgery}, vol.~13, no.~3,
  2017.

\bibitem{szegedy2015going}
C.~Szegedy, W.~Liu, Y.~Jia, P.~Sermanet, S.~Reed, D.~Anguelov, D.~Erhan,
  V.~Vanhoucke, A.~Rabinovich \emph{et~al.}, ``Going deeper with
  convolutions.''\hskip 1em plus 0.5em minus 0.4em\relax CVPR, 2015.

\bibitem{chung2014empirical}
J.~Chung, C.~Gulcehre, K.~Cho, and Y.~Bengio, ``Empirical evaluation of gated
  recurrent neural networks on sequence modeling,'' \emph{arXiv preprint
  arXiv:1412.3555}, 2014.

\bibitem{ioffe2015batch}
S.~Ioffe and C.~Szegedy, ``Batch normalization: Accelerating deep network
  training by reducing internal covariate shift,'' in \emph{International
  Conference on Machine Learning}, 2015, pp. 448--456.

\bibitem{nair2010relu}
V.~Nair and G.~E. Hinton, ``Rectified linear units improve restricted boltzmann
  machines,'' in \emph{Proceedings of the 27th International Conference on
  Machine Learning (ICML-10)}, 2010, pp. 807--814.

\bibitem{srivastava2014dropout}
N.~Srivastava, G.~Hinton, A.~Krizhevsky, I.~Sutskever, and R.~Salakhutdinov,
  ``Dropout: A simple way to prevent neural networks from overfitting,''
  \emph{The Journal of Machine Learning Research}, vol.~15, no.~1, pp.
  1929--1958, 2014.

\bibitem{gao2014JIGSAW}
Y.~Gao, S.~S. Vedula, C.~E. Reiley, N.~Ahmidi, B.~Varadarajan, H.~C. Lin,
  L.~Tao, L.~Zappella, B.~B{\'e}jar, D.~D. Yuh \emph{et~al.}, ``Jhu-isi gesture
  and skill assessment working set (jigsaws): A surgical activity dataset for
  human motion modeling,'' in \emph{MICCAI Workshop: M2CAI}, vol.~3, 2014.

\bibitem{2015DL_review}
J.~Schmidhuber, ``Deep learning in neural networks: An overview,'' \emph{Neural
  Networks}, vol.~61, pp. 85--117, 2015.

\bibitem{kingma2014adam}
D.~Kingma and J.~Ba, ``Adam: A method for stochastic optimization,''
  \emph{arXiv preprint arXiv:1412.6980}, 2014.

\end{thebibliography}


\end{document}